# Deep Reinforcement Learning-based Obstacle Avoidance for Robot Movement in Warehouse Environments

Keqin Li[1,a*]
Department of Computer Science
AMA University
Quezon, Philippines
[a]Email: keqin157@gmail.com

Jiajing Chen[1,b*],
Courant Institute
New York University
251 Mercer St, New York, NY 10012, USA
[b]Email: jc12020@nyu.edu

Denzhi Yu[2,c]
School of Information
University of California, Berkeley
University Avenue and, Oxford St, Berkeley, CA 94720, USA
[c]Email:dezhi.yu@berkeley.edu

Tao Dajun[3,d]
School of Engineering
Carnegie Mellon University
5000 Forbes Avenue, Pittsburgh, PA, USA
[d]Email:dajunt@alumni.cmu.edu

Xinyu Qiu[4,e]
College of Engineering
Northeastern University
Seattle, USA
[e]Email:qiu.xiny@northeastern.edu

Lian Jieting[5,f]
New York University
School of Professional Studies
50 West 4th Street, New York, NY, USA
[f]Email:jl14282@nyu.edu

Sun Baiwei[6,g]
School of Information and Computer Sciences
University of California, Irvine
510 Aldrich Hall Irvine, CA, USA
[g]Email:sunbaiwei123@gmail.com

Zhang Shengyuan[7,h]
College of Computing and Information Science
Cornell University
616 Thurston Ave. Ithaca, NY, USA
[h]Email:sophiazhang217@gmail.com

Zhenyu Wan[8,i]
independent researcher
2513 Las Palmas Ln, Plano, TX 75075,USA
[i]Email:dd.othse@gmail.com

Ran Ji[9,j]
College of Engineering
Cornell University
616 Thurston Ave, Ithaca, NY, 14853, USA
[j]Email:rj369@cornell.edu

Bo Hong[10,k]
School of Informatics, Computing, and Cyber Systems
Northern Arizona University
Flagstaff, AZ, USA
[k]Email:hongbo2904@gmail.com

Fanghao Ni[11,l]
School of Informatics, Computing, and Cyber Systems
Northern Arizona University
Flagstaff, AZ, USA
[l]Email:fn232@nau.edu

*Corresponding author:*keqin157@gmail.com

* jc12020@nyu.edu

• Authors 1 [a*] and 1 [b*] contributed equally to this paper and are co-first authors.

**Abstract:** At present, in most warehouse environments, the accumulation of goods is complex, and the management personnel in the control of goods at the same time with the warehouse mobile robot trajectory interaction, the traditional mobile robot can not be very good on the goods and pedestrians to feed back the correct obstacle avoidance strategy, in order to control the mobile robot in the warehouse environment efficiently and friendly to complete the obstacle avoidance task, this paper proposes a deep reinforcement learning based on the warehouse environment, the mobile robot obstacle avoidance Algorithm. Firstly, for the insufficient learning ability of the value function network in the deep reinforcement learning algorithm, the value function network is improved based on the pedestrian interaction, the interaction information between pedestrians is extracted through the pedestrian angle grid, and the temporal features of individual pedestrians are extracted through the attention mechanism, so that we can learn to obtain the relative importance of the current state and the historical trajectory state as well as the joint impact on the robot's obstacle avoidance strategy, which provides an opportunity for the learning of multi-layer perceptual machines afterwards. Secondly, the reward function of reinforcement learning is designed based on the spatial behaviour of pedestrians, and the robot is punished for the state where the angle changes too much, so as to achieve the requirement of comfortable obstacle avoidance; Finally, the feasibility and effectiveness of the deep reinforcement learning-based mobile robot obstacle avoidance algorithm in the warehouse environment in the complex environment of the warehouse are verified through simulation experiments.

## Introduction

Warehouse robots work in complex and variable scenarios covering complex stacked goods and warehouse personnel. The traditional obstacle avoidance algorithm uses the information of the surrounding obstacles at the current moment and relies on methods such as geometric configuration or sampling to obtain the final obstacle avoidance strategy[1,2], which does not take into account the changes of the future state of the surrounding pedestrians, and in the complex environment, the algorithm is prone to unnatural behaviours, such as oscillation, due to the randomness of the pedestrian's movement, which makes the robot's control signals jump frequently, and fails to satisfy the requirements of safety and comfort in warehouse environments. . Therefore, in order to efficiently complete the obstacle avoidance task in the cargo-intensive warehouse environment, the obstacle avoidance algorithm of the mobile robot needs to analyse the pedestrian's behaviour. At the same time, with the development of human-computer interaction concept, the robot's obstacle avoidance task not only needs to meet the high efficiency and real-time performance, but also needs to take into account the comfort requirements of pedestrians in the process of interaction with people.

For pedestrian behaviour, the historical trajectories of pedestrians can be fitted by mathematical models such as Hidden Markov Models and Gaussian Mixture Models to analyse their trajectories, but these methods are poorly adapted and do not take into account the physical environment constraints, the influence of the surrounding intelligences, and social norms and other specific factors of the intelligences' travels, so that the pedestrian trajectories can not be fitted and predicted very well. Therefore, many scholars have established pedestrian flow models to better analyse pedestrian behaviour by studying pedestrian movement and its characteristics. Among them, the social force model proposed by Helbing and Molnar is one of the most widely used traditional pedestrian movement models, which describes the relationship between pedestrians and the surrounding obstacles as well as their own target points in the form of force, and assumes that the potential repulsive force between the pedestrians is a monotonically decreasing function of the distance between the two and the pedestrian's occupancy radius, which is of some practical significance. The disadvantage is that this type of method is relatively rough, and needs to be adjusted according to the environment of the parameters, there are certain limitations. With the development of time-series network, the time-series network model based on large-scale pedestrian movement data set can predict the trajectory of pedestrian movement more accurately. Among them, Social-LSTM (Social Long Short Term Memory)and Social-GAN (Social Generative Adversarial Network) firstly rasterise the surrounding environment to get the local map information of pedestrians[3,4], and use the network to learn the potential rules between pedestrians and predict the trajectories of pedestrians. Using the network to learn the potential rules between pedestrians, accurate trajectory information is predicted, but the discrete processing of rasterisation loses a certain amount of environmental information, and due to the large amount of computation of the prediction model itself and the re-planning problem caused by the existence of prediction errors, it is difficult for obstacle avoidance algorithms based on the prediction of pedestrian trajectory to meet the requirements of real-time. In addition, the obstacle avoidance algorithm based on the trajectory prediction can easily cause the robot to lose its

In addition, the trajectory prediction-based obstacle avoidance algorithm can easily lead to the freezing phenomenon of the robot in a crowded situation, thus failing to find a feasible path for the robot.

Google proposed deep reinforcement learning algorithm in 2015 [5], which combines the perception ability of deep learning and the decision-making ability of reinforcement learning to better solve the perception decision-making problem of complex systems, so many scholars apply it in the learning of robot obstacle avoidance strategy [6]. In crowded environments, the obstacle avoidance algorithm based on deep reinforcement learning integrates pedestrian behaviour prediction and robot motion planning, and the control output signal of the robot is obtained directly from the environment state, which ensures the real-time and reliability of the obstacle avoidance algorithm, and it has become a new hotspot of research.Everett et al [7] use a multilayer perceptron to fit the value function that Through the continuous interaction between the intelligent body and the environment, the value function network can learn the uncertainty of pedestrian movement to a certain extent, which improves the obstacle avoidance efficiency of the robot in the pedestrian-intensive environment. However, the obstacle avoidance algorithm based on the deep reinforcement learning algorithm still has unnatural obstacle avoidance behaviours due to the strong randomness of pedestrian behaviours and the limited learning ability of the value function network. The comfort requirement mainly takes into account that the robot needs to keep a certain distance from people, which not only avoids collision, but also prevents pedestrians from emotional discomfort caused by being too close to the robot . The personal space theory proposed by Hall [8] is widely used in human-robot interaction, which classifies the distance between people into four types: intimate distance, private distance, social distance and public distance, and in order to satisfy the comfort requirements of human-robot interaction, the robot should avoid appearing in intimate or private distance.Pande [9] modified the path planning algorithm based on the social requirements such as pedestrian comfort. The algorithm was modified to make the obstacle avoidance strategy comfortable, but it is less adaptable as it introduces a large number of parameters as well as a complex scenario design, which needs to be adjusted after the environment changes and may require redesigning the obstacle avoidance algorithm for the new environment.

In order to learn pedestrian behaviour more accurately and to make the obstacle avoidance algorithm meet the comfort requirements without introducing additional computational parameters, this paper proposes a deep reinforcement learning-based obstacle avoidance algorithm for mobile robots in crowded environments. The algorithm is inspired by the pedestrian trajectory prediction method, and improves the learning ability of the value function in the deep reinforcement learning algorithm by analysing the behaviour of pedestrians in the environment; by using the reward function to guide the intelligent body appropriately, the obstacle avoidance strategy that meets the requirements of human-robot interaction can be obtained without accurate modelling, so as to achieve a balance between the robot's obstacle avoidance efficiency and the pedestrian's comfort level. The main innovations of this paper are.

1) Dynamic obstacles around pedestrians are encoded using angular pedestrian grid and temporal features of pedestrians are extracted using attention mechanism, which gives richer information about pedestrian dynamics and improves the learning ability of the value function network.

2) Based on the spatial behaviours of the pedestrians, the reward function for reinforcement learning is redesigned by penalising the state of entering into the pedestrian within a private distance as well as the state of excessive angular velocity change This makes the robot achieve a balance between pedestrian comfort and obstacle avoidance efficiency in the process of obstacle avoidance, so that it can better interact with people.

# 1 Obstacle avoidance algorithm based on deep reinforcement learning

The goal of using reinforcement learning algorithm to solve the obstacle avoidance problem is to obtain the optimal strategy π* from the joint state $s^{jn}_t$ of the robot to the control output at, as shown in Fig. 1. In the obstacle avoidance algorithm based on reinforcement learning, the intelligent body obtains the state $s^{jn}_t$ of the robot and all the pedestrians in the surrounding area at the current moment by interacting with the environment, and then discretises the robot's control signal into a certain size of action space by one-step prediction.

Then the control signals of the robot are discretised into an action space of a certain size, and the state after the execution of each control signal in the action space is predicted by a one-step prediction model, and the predicted state $s^{jn}_{t+\Delta t}$ is obtained; then the predicted states $s^{jn}_{t+\Delta t}$ is inputted into the value function network, and the value of the state is obtained by combining the reward function, and the optimal action in the action space corresponding to the maximal value is selected as the final output action of the robot, which is $a_t$, as shown in Equation (1). During the interaction between the intelligent body and the environment, the value function network V* is continuously optimised until it converges.

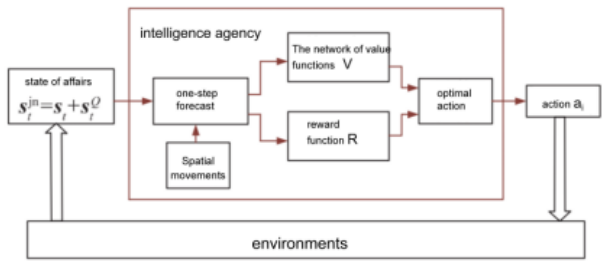

**Fig. 1** Obstacle avoidance algorithm based on deep reinforcement learning

$$a_t = \arg\max R(S^{jn}_t, a_t) + \gamma^{\Delta t \bullet v} prefV^*(S^{jn}_{t+\Delta t}) \quad (1)$$

where $s^{jn}_{t+\Delta t}$ = prop( $s^{jn}_t$, Δt, $a_t$ ), denotes that the prediction model approximates the joint state of the robot after the Δt moment by predicting the motion of the robot and its surrounding pedestrians after the Δt moment through a simple linear model. The following is the specific selection method for each element of reinforcement learning:

State Space: Considering that both the robot's own state and the state of other pedestrians in the environment will affect the final decision, the joint state $s^{jn}_t = [s_t, s^o_t]$ is used as the input state for reinforcement learning at time t. The input state is the robot's own state and the state of other pedestrians in the environment. Where, $s_t = [p_x, p_y, v_x, v_y, r, g_x, g_y, v_{pref}, \phi]$ is the state of the robot, which includes the robot's position, movement speed, size, target point, optimal running speed and steering angle; sot = $[s^{1o}_t, s^{2o}_t, ..., s^{no}_t]$ is the state of all the pedestrians in the environment, which is different from st, and sot = $[s^{1o}_t, s^{2o}_t, ..., snot]$ is the state of all pedestrians in the environment. Since in a crowded environment, the robot cannot obtain the information of pedestrians' target point, optimal running speed and steering angle, the state siot of the ith pedestrian includes only the pedestrian's position, movement speed, size and the coded information of the surrounding pedestrians, $s^{io}_t = [p_x, p_y, v_x, v_y, r, r_i]$.

Action space: The action space is taken as a set of feasible robot velocity control signals, the robot's action at time t $a_t$ = [v, ω], where v, ω represent the linear and angular velocities of the robot, respectively.

Value Functions: The goal of the reinforcement learning algorithm is to obtain an optimal value function $V^*$, and the optimal policy π* can be obtained through a network of value functions, so that the robot receives the optimal desired reward in its interaction with the environment.

$$V^*(S^{jn}_t) = \sum_{t'=t}^{T} \gamma^{t' \bullet v_{pref}} R_{t'}(s^{jn}_{t'}, \Pi^*(s^{jn}_{t'}))$$

$$\pi^*(s^{jn}_t) = \arg\max R(s^{jn}_t, a_t) + \gamma^{\Delta t \bullet V_{pref}} \int_{s^{jn}_{t+\Delta t}} p(s^{jn}_t, a_t, s^{jn}_{t+\Delta t}) V^*(s^{jn}_{t+\Delta t}) ds^{jn}_{t+\Delta t}$$

(2)

where R( $s^{jn}_t$, $a_t$ ) is the reward value at time t; γ ∈ ( 0, 1) is the decay coefficient; P($s^{jn}_t$, $a_t$, $s^{jn}_{t+\Delta t}$) is the state transfer probability; and vpref denotes the speed of the robot running when there are no obstacles, which serves as a normalisation term in the decay coefficient afterwards.

The value function can learn certain information about the environment through the continuous interaction between the robot and the environment, but the complex pedestrian interaction rules cannot be learnt well using only a shallow multilayer perceptron, in this paper, the value function is modified in order to improve the learning ability of the value function in crowded and complex environments, and further modifications are made to the reward function to address the requirements of pedestrian comfort. In addition, this paper uses an LSTM network for the case of variable number of pedestrians.

## 2 Specific algorithms

In this section, by introducing pedestrian interaction information in the value function network and modifying the reward function for reinforcement learning, obstacle avoidance algorithms that meet the requirements of human-computer interaction are obtained based on deep reinforcement learning.

### 2.1 Improvement of value function network based on pedestrian interaction

As shown in Figure 2, the improved value function network μ is composed of three parts: the pedestrian interaction information module (crowd interaction module), the $L_{STM}$ network $\Phi_{LSTM}(\cdot)$, and the multilayer perceptron (MLP). The pedestrian interaction module first extracts pedestrian features from the original agent state sjnt. Then, through the $L_{STM}$ network $\Phi_{LSTM}(\cdot)$, the pedestrian features of an indefinite number are combined to obtain the joint hidden state ho of all pedestrians. Finally, ho and the robot's own state are jointly input into the multilayer perceptron network $\psi_M(\cdot)$ to obtain the corresponding value. This section mainly introduces the specific method for analyzing pedestrian interaction behavior in the improved value function network.

The decisions between pedestrians influence each other. If the relationship between each pair of pedestrians is accurately described, it will lead to a time complexity of $O(N^2)$. As the number of pedestrians in the environment increases, the computational burden will become extremely heavy. In addition, pedestrians who are far apart will hardly have an impact on each other's movements. Describing the relationship between the two will not only increase the computational complexity but also increase the learning burden of the network. Accordingly, this paper adopts a special hybrid grid - the angular pedestrian grid (APG)to encode the local environment of pedestrians to eliminate information that is useless for obstacle avoidance strategies. In order to better capture the dynamic characteristics of pedestrians, the angular pedestrian grid (APG) introduces a high-resolution grid, increasing the extensiveness of information. Figure 3 shows the encoding process for the i-th pedestrian $P^i_H$. Let the total number of pedestrians be N. The angular pedestrian grid (APG) only considers pedestrians within a circle centered on pedestrian $P^i_H$ and with a radius of rmax. The circle is equally divided into K sectoral areas, and the nearest distance to pedestrian $P^i_H$ in each sectoral area is selected as the encoding value of that sectoral area. After being encoded by the angular pedestrian grid (APG), pedestrian $P^i_H$ is output as a one-dimensional vector $r_i$ = [$r_i$,1,...,$r_i$,k,...,$r_i$,K], where the mathematical representation method of ri,k (the encoding value of the k-th sector) is shown in Equation (3).

$$r_{i,k} := \min(r_{\max} . \min(\{\rho_{i,j}, j \in N_{i,k}\}))$$

$$N_{i,k:} = \{j \in \{1,...,N\} \setminus \{i\}, \vartheta_{i,j} \in [\gamma_k, \gamma_{k+1}]\} \quad (3)$$

Where k ∈ {1,..., K}, and the sector angle γk corresponding to each grid is γk := (k−1)2k/K. ($\rho_{i,j}$, $\varphi_{i,j}$) represents the coordinates of pedestrian $P^j_H$ in the polar coordinate system centered on pedestrian $P^i_H$. Therefore, $r_i$ is only related to the pedestrian positions.

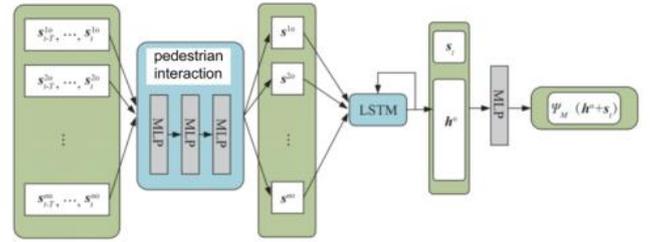

**Fig. 2** Improved network structure of value function

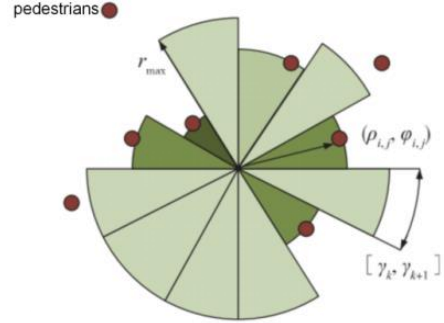

**Fig. 3** Angular pedestrian grid

Compared with the standard 2D grid, APG only linearly affects the dimension of the input, but it can still capture changes in radial distance with continuous resolution rather than discrete grid cells. In addition, the closer pedestrian $P^j_H$ is to pedestrian $P^i_H$, the more accurate the observability of the angular position change of pedestrian $P^j_H$ becomes. For an APG grid with given parameters, the state soit = [$p^i_x$, $p^i_y$, $v^i_x$, $v^i_y$, r, $r_i$] of pedestrian $P^i_H$ at time t can be obtained. This state variable will be used as the input of the improved value function network.

The movement of the robot is not only influenced by the pedestrian information at the current moment, but also has a great relationship with the pedestrian information at previous moments. Moreover, by relying on the joint influence of multiple moments, the short-sightedness of the obstacle avoidance strategy can be overcome, and the problem of control signal jumps caused by relying only on the immediate environmental information at the current moment can be reduced. Historical data can be used for obstacle avoidance to achieve better obstacle avoidance effects. In time series prediction, the attention mechanism has achieved good application results. In this paper, the self-attention mechanism is used to extract the time series features of a single pedestrian, and learn the relative importance of T moments of data including the current moment and the joint influence on the robot's obstacle avoidance strategy.

The continuous trajectory sequence $[s^{oi}_{t-T},..., s^{oi}_{t-k},..., s^{oi}_t]$ of pedestrian $P^i_H$ within time T is known. As shown in Figure 4, first, the pedestrian features at each moment are encoded, and a fixed-length output $e_m$ is obtained through the multilayer perceptron $\psi_e(\cdot)$ (MLP).

$$e_m = \phi_e(s^{io}_{t-m}, W_e) \quad (4)$$

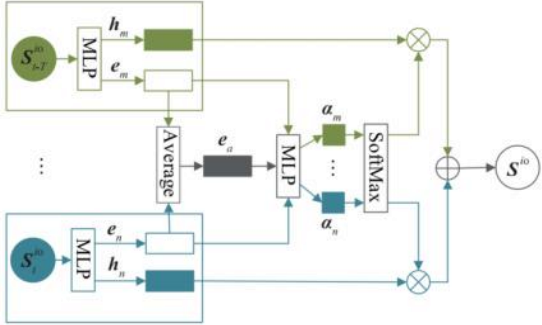

Fig. 4 Attention mechanism model

Then, the vector $e_m$ is input into the perceptron $\psi_h(\cdot)$, and the interaction characteristic vector $h_m$ between the states of pedestrian $P^i_H$ at different times is extracted from it.

$$e_k = \psi_h(e_m, W_h) \quad (5)$$

Next, through the multilayer perceptron $\psi_\alpha(\cdot)$, the relative importance of pedestrian states at different times is calculated, and the attention score $\alpha_m$ of each interaction feature vector is obtained.

$$e_a = \frac{1}{T}\sum_{k=1}^{T} e_k \quad (6)$$

$$a_m = \psi_\alpha(e_m, e_a, W_\alpha) \quad (7)$$

$e_a$ is the average pooling vector of all moments. $\psi_e(\cdot)$, $\psi_h(\cdot)$, and $\psi_\alpha(\cdot)$ are fully connected networks. Their nonlinear activation functions are all Re-LU. $W_e$, $W_h$, and $W_\alpha$ are the weight values of the fully connected networks respectively. The specific network parameters will be given in the experimental section. Using the interaction feature vector $e_m$ and the corresponding attention score $\alpha_m$ at each moment, the final state output $e^o$ of pedestrian $P^j_H$ can be obtained by linearly combining the interaction feature vectors of all moments.

$$e^o = \sum_{k=1}^{T} soft\max(a_k)e_m \quad (8)$$

Finally, the LSTM network processes an indefinite number of obstacles in the environment and finally obtains a fixed-length output vector $h^o$.

## 2.2 Comfort reward function

In the scenario where the robot interacts with people, if pedestrians do not initiate service requests to the robot, in order to ensure the comfort requirements of pedestrians, the robot should try to avoid appearing within the comfort range of pedestrians as much as possible, that is, the robot should not enter the personal distance of pedestrians. In addition, according to reference, the lateral comfortable distance for the robot to overtake pedestrians from behind to the front is 0.7 m. If it is less than this distance, it will also cause discomfort to pedestrians. Therefore, based on the above requirements for pedestrian comfort, this paper modifies the reward function and penalizes the state where the robot enters the comfort range of pedestrians.

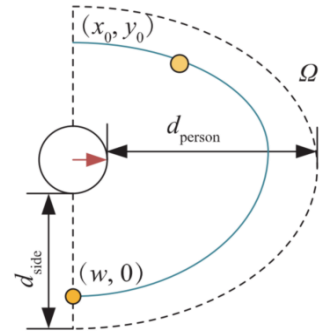

Fig. 5 Personal distance zones for people

According to the knowledge of pedestrian comfort, the pedestrian comfort zone $\Omega$ is defined: taking the pedestrian as the coordinate origin and the forward direction as the x-axis, a pedestrian coordinate system is established. The pedestrian comfort zone is represented by a semi-ellipse Ω with the pedestrian as the origin, the maximum personal distance $d_{person}$ as the major axis, and the side comfort distance $d_{side}$ as the minor axis (only considering the situation that appears within the pedestrian's field of vision. $d_{side}$ is 0.7 m and $d_{person}$ is 1.2 m).Suppose the coordinates of the robot in the pedestrian coordinate system are $(x_0, y_0)$. The indicator function $I(x_0, y_0)$ can be used to determine whether the robot has driven into the pedestrian comfort zone Ω. $I(x_0, y_0)$ being 1 means the robot has entered the pedestrian comfort zone, and $I(x_0, y_0)$ being 0 means the robot has not entered the pedestrian comfort zone.

$$I(x_0, y_0) = 0, \frac{x_0^2}{d_{person}^2} + \frac{y_0^2}{d_{side}^2} \geq 1$$

$$I(x_0, y_0) = 1, \frac{x_0^2}{d_{person}^2} + \frac{y_0^2}{d_{side}^2} \leq 1 \quad (9)$$

If the robot reaches the pedestrian comfort zone of pedestrians, that is, $I(x_0, y_0) = 1$, it will cause discomfort to pedestrians, and the robot needs to be given corresponding penalties. The magnitude of the penalty depends on the position of the robot in the pedestrian comfort zone. The closer the robot is to the pedestrian, the greater the penalty. As shown in Figure 5, the robot is marked in yellow. If the robot (yellow circle) reaches the pedestrian comfort zone, it can be assumed that there is the same penalty on the ellipse with the same ratio of major and minor axes as the pedestrian comfort zone at the robot's coordinate in the pedestrian comfort zone. That is, all points on the blue ellipse shown in Figure 5 have the same penalty term, so $(x_0, y_0)$ and $(\omega, 0)$ have the same penalty. Since being closer to people is more likely to cause people's discomfort, and the discomfort will become greater and greater. Therefore, an exponential function is used to characterize this penalty value. The maximum value of the penalty is $r_p$. So the obtained penalty function related to $\omega$ is as follows:

$$R_t^{social}(\omega) = -r_p(e^{-(\omega - d_{side})} - 1) \quad (10)$$

where w is the short axis of the blue ellipse.

$$w = \sqrt{\frac{(x_0^2 \cdot d_{side}^2)}{d_{person}^2 + y_0^2}} \quad (11)$$

The original reward function $R^o_t(s^{jn}_t, a_t)$ is the reward function used in reference, in which the penalty term when the robot is close to an obstacle is removed.

$$R_t^o(s_t^{jn}, a_t) = -0.25, if, d_t \leq 0$$

$$R_t^o(s_t^{jn}, a_t) = 2, elseif, p_t = p_g$$

$$R_t^o(s_t^{jn}, a_t) = 0, otherwise \quad (12)$$

At the same time, in order to avoid excessive changes in the angular velocity of the robot, a penalty is imposed on cases where the angular velocity changes too much.

$$R_t^A(s_t^{jn}, a_t) = -r_{Angle}, if, \varphi_t - \varphi_{t-1} \geq \pi$$

$$R_t^A(s_t^{jn}, a_t) = 0, \quad otherwise \quad (13)$$

The final reward function is

$$R_t(s_t^{jn}, a_t) = R_T^0(s_t^{jn}, a_t) + R_t^{social}(s_t^{jn}, a_t) + R_t^{jn}(s_t^{jn}, a_t) \quad (14)$$

Through this reward function, the obstacle avoidance strategy of the robot can be more in line with human comfort criteria.

### 2.3 Network training

The training method of the entire reinforcement learning algorithm adopts the temporal difference method. To increase the utilization rate of samples, the information of each interaction between reinforcement learning and the environment is stored in the replay buffer. During network training, training data is obtained by sampling from the replay buffer through the method of uniform sampling. The fixed target network method is used to accelerate the convergence of the algorithm and reduce the variance of the algorithm. At the same time, to further accelerate convergence, before training, the dataset generated by the traditional method ORCA is used for network pre-training.

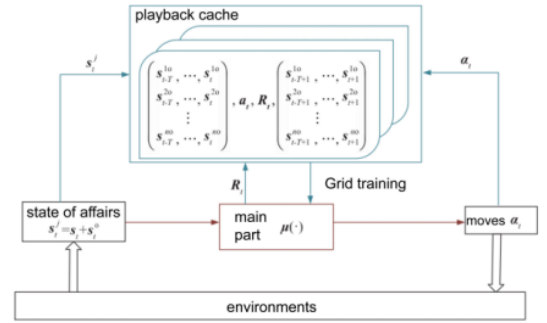

Fig. 6 Network training

# 3 Experimental validation

This chapter will verify the effectiveness and generalization of the improved value function network and the feasibility of the comfort reward function by comparing the performance of obstacle avoidance algorithms with different encoding methods and the performance of obstacle avoidance algorithms before and after modifying the reward function.

Hardware platform: The hardware platform for the experiment is a desktop computer with Intel Core i7-7700 3.60 GHz and NVIDIA Geforce GTX 1050 Ti.

Simulation environment: All experiments are carried out in a simulation environment built based on Gym. In the simulation environment, a circle with a radius of 0.3 m is used to replace pedestrians and robots. The task of pedestrians and robots is to run from the starting point to the target point. Among them, the movement of pedestrians is controlled by the traditional obstacle avoidance algorithm ORCA strategy. The selection of starting points and target points adopts a cross-encounter method. That is, the initial positions of pedestrians and robots are randomly located on the same circle with a radius of 4 m, and the target point is the point that is centrally symmetric to its initial position with respect to the center of the circle.

Robot action space setting: The angular velocity and linear velocity of the robot in the simulation. In the experiment, the discrete action space of the robot is selected as the combination of the robot's linear velocity space and angular velocity space, with a size of 80. Among them, the linear velocity space is uniformly sampled in [0, $v_{pref}$], and the angular velocity space is uniformly sampled in [0, $2\pi$], with sampling sizes of 5 and 16 respectively.

Network setting: This paper first collects data from 3,000 training rounds through the traditional method ORCA to complete the initialization of network weights. After the pre-training is completed, in order to balance the exploration of unknown states and the utilization of existing results, the e-greedy method is used for training. At the beginning of training, in order to explore the unknown states in the environment more effectively, the exploration rate is set to 0.5. Then, as the number of training increases, the exploration rate is continuously reduced. At 5,000 rounds, it is reduced to 0.1. Subsequent training will no longer change the exploration rate. The remaining hyperparameters and the network parameters of the value function during the experiment are shown in Tables 1 and 2 respectively.

Tab. 1 Hyper parameters setting

| parametric | pre-training | training |
|---|---|---|
| N | 6 | 6 |
| $V_{pref}$ | 1 | 1 |
| K | 12 | 12 |
| $r_{max}$ | 3 | 3 |
| T | 5 | 5 |
| learning rate | 0.01 | 0.001 |
| batch-size | 100 | 100 |
| $\Delta t$ | 0.25 | 0.25 |
| $\gamma$ | 0.9 | 0.9 |

Tab. 2 Parameters of networ

| lattices | Mesh Parameter Setting |
|---|---|
| $\Psi_e(\cdot)$ | [150,100] |
| $\Psi_h(\cdot)$ | [150,50] |
| $\Psi_a(\cdot)$ | [100,100,1] |
| $\Psi_{LSTM}(\cdot)$ | [50,50] |
| $\Psi_M(\cdot)$ | [150,100,100,1] |

## 3.1 Experimental validation of the improved value function network

### 3.1.1 Feasibility experiment verification

This subsection verifies the effectiveness of the improved value function network in this paper by comparing it with ORCA, CADRL, and obstacle avoidance algorithms with two different encoding methods based on social force and local map. In order to avoid the influence of the comfort reward function on this module, the reward function here adopts the reward function in paper.Therefore, the difference between the obstacle avoidance algorithms based on reinforcement learning lies only in the processing method of pedestrian interaction information. For convenience of expression, the algorithm based on social force encoding is called SF_RL, the method based on local map encoding is called LM_RL, and the algorithm based on APG encoding proposed in this paper is called APG_RL.

Figure 7 shows the change curves of the cumulative return function after 10,000 rounds of training of four reinforcement learning-based obstacle avoidance algorithms, CADRL, SF_RL, LM_RL, and APG_RL, in a simulation environment with 6 pedestrians. Among them, the convergence performance of CADRL, LM_RL, and APG_RL is significantly higher than that of SF_RL. This is because there may be errors in the description of the interaction information between pedestrians using the social force model, thus reducing the convergence of the network.

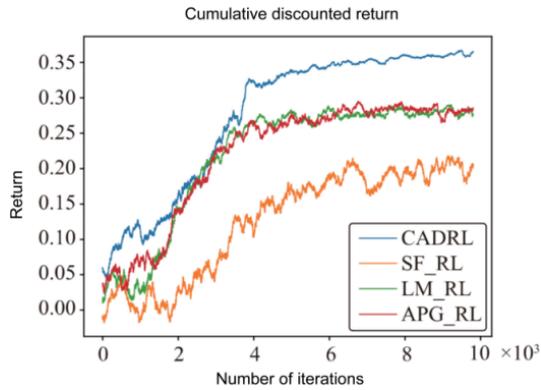

**Fig．7** Learning return curve of different algorithms

After the training is completed, 500 tests are respectively conducted on the above five algorithms in a simulation environment with 6 pedestrians. The test results are shown in Table 3. Among them, the reinforcement learning-based algorithm uses a multilayer perceptron to fit the value function, and through continuous interaction between the robot and the environment, the interaction characteristics between pedestrians are learned to a certain extent, which improves the obstacle avoidance performance of the robot in a pedestrian-dense environment. The success rate in the test process is higher than that of the traditional ORCA algorithm. However, due to different ways of extracting pedestrian interaction information, the performance of the algorithms also differs. Since CADRL does not process pedestrian interaction features, the success rate of obstacle avoidance in the test process is low, indicating that only using a shallow network cannot extract information in a complex dynamic environment well. While SF_RL, LM_RL, and APG_RL, due to the addition of the extraction of pedestrian interaction features and the temporal features of pedestrians, have certain prior knowledge, which greatly improves the success rate of obstacle avoidance. Compared with SF_RL, the obstacle avoidance algorithms LM_RL and APG_RL using local information encoding have better convergence in the network training process, and the obstacle avoidance time is shorter and the efficiency is higher.

**Tab．3** Quantitative results comparison among different methods

| arithmetic | success rate | failure rate | average time/s | return |
|---|---|---|---|---|
| ORCA | 0.33 | 0.66 | 11.03 | -0.0652 |
| CADRL | 0.59 | 0.40 | 10.68 | 0.0738 |
| SF_CADRL | 0.96 | 0.03 | 11.86 | 0.2608 |
| LM_RL | 0.99 | 0.01 | 10.97 | 0.3190 |
| APG_RL | 0.99 | 0.01 | 10.81 | 0.3196 |

Figure 8 shows the obstacle avoidance results of different obstacle avoidance algorithms in the same test scenario. The yellow solid circle represents the robot, and the circles of other colors represent different pedestrians. The numbers in the circles represent the current running time of the robot. In Figure 8(a), the robot control algorithm uses ORCA. The robot collides with pedestrians at 3.5 s, resulting in navigation failure. The other methods reach the target point at 9.8 s, 10.2 s, 9.2 s, and 9.0 s respectively. Among them, APG_RL reaches the target point in the shortest time. Combined with Table 3, it can be seen that the average navigation time of APG_RL after 500 tests is only 10.82 s, which greatly improves the efficiency of obstacle avoidance and well verifies the effectiveness of the improved value function network in this paper.

In addition to higher navigation efficiency, through experiments, it is found that APG_RL can adapt to some more complex environments and still complete the navigation task when LM_RL fails in obstacle avoidance. The specific trajectory diagram is shown in Figure 9. As shown in Figure 9(a), six pedestrians drive from different directions towards the center. At the 4th second, the robot controlled by the LM_RL algorithm collides with pedestrians because it drives into the middle of pedestrians. While APG_RL encodes the surrounding pedestrians by adopting the angular pedestrian grid and obtains continuous values of the surrounding pedestrians. Compared with the gridded processing of the local map, it can capture the changes of pedestrians more accurately. Therefore, in the complex environment shown in Figure 9, when the robot is close to pedestrians, it changes its direction of movement and avoids entering the crowd. As shown in Figure 9(b), the robot turns right at 4 seconds and finally successfully reaches the target point.

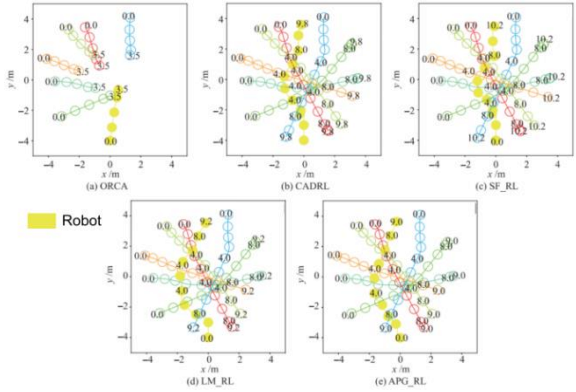

Fig．8 Comparison of robot trajectories with different algorithms under the same simulatio condition

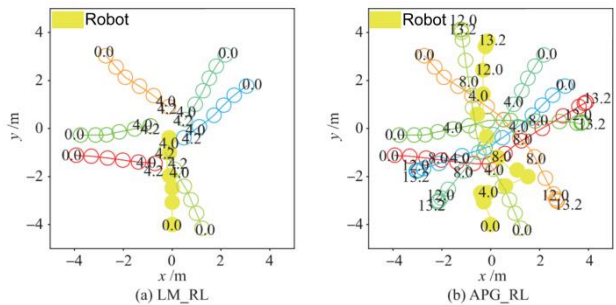

Fig．9 Trajectory comparison in a test case between LM_RL and APG_RL

### 3.1.2 Experimental verification of generalizability

The number of pedestrians in the simulation environment during the training of the APG_RL algorithm is 6. In order to further test the generalization performance of the model, without retraining the network, the number N of pedestrians in the simulation environment is changed. Table 4 shows the results of 500 obstacle avoidance tests with different numbers of pedestrians. The experimental results show that when the number of pedestrians changes, the success rate of robot obstacle avoidance remains above 0.91. This experiment fully proves that the APG_RL algorithm can deal with the changes of pedestrians in the environment well. Especially in the simulation environment with 8 pedestrians, it can still achieve a high success rate, so it can better deal with complex pedestrian environments.

Tab．4 Simulation results of different numbers of pedestrians using APG_RL

| Number of pedestrians | 3 | 4 | 5 | 6 | 7 | 8 |
|---|---|---|---|---|---|---|
| success rate | 1.00 | 1.00 | 0.99 | 0.99 | 0.98 | 0.91 |
| average time | 10.20 | 10.18 | 10.81 | 10.82 | 12.30 | 12.75 |

### 3.2 Experimental verification of the pedestrian comfort reward function

This experiment experimentally verified the feasibility of the reward function proposed in this paper. The values of $r_p$ and $r_{Angle}$ are both 0.02. Without changing the APG_RL obstacle avoidance algorithm, training is carried out according to the comfort reward function designed by formula (13), and the comfortable obstacle avoidance strategies APG_CRL and APG_CARL are obtained respectively. The reward function of APG_CRL only adds the $R^{social}_t(w)$ term, and the reward function of APG_CARL adds both Rsocialt(w) and RA t($s^{jn}_t$, $a_t$). Table 5 shows the results obtained by testing the above three algorithms 500 times in a simulation environment with 6 pedestrians.

Tab．5 Simulation test results of different reward function using APG_RL

| arithmetic | success rate | failure rate | average time | gap |
|---|---|---|---|---|
| APG_RL | 0.99 | 0.01 | 10.82 | 0.14 |
| APG_CRL | 0.99 | 0.01 | 10.96 | 0.64 |
| APG_CARL | 0.99 | 0.01 | 10.77 | 0.50 |

### 3.2.1 Experimental verification of the reward function for pedestrian comfort

As can be seen from Table 5, both APG_RL and APG_CRL have a high navigation success rate. While the average time for the robot to reach the target point only increases by 0.14 s, the closest distance to pedestrians increases from 0.14 m to 0.64 m, which well balances the requirements of navigation efficiency and pedestrian comfort.

In addition, the pedestrian comfort reward function not only increases the maximum distance from pedestrians during navigation. Compared with the APG_RL obstacle avoidance algorithm, APG_CRL can be well applied to some more complex environments because it penalizes the state of entering the pedestrian's personal distance. In the dynamic and complex environment shown in Figure 10(a), since APG_RL only penalizes the robot when it is close to pedestrians, the robot will drive towards the place where people gather. While APG_CRL penalizes the state where the robot enters the pedestrian comfort range, making the robot tend to choose the behavior of keeping a relatively long distance from people. As shown in Figure 10(b), the robot will wait in place for pedestrians to leave before driving towards the target point.

### 3.2.2 Experimental verification of the angular velocity reward function

Excessive angular changes are not only unfavorable for the control of actual robots, the acquisition of sensor data, etc., but also have an impact on pedestrians in the surrounding environment. Therefore, APG_CARL penalizes cases where the robot's angular change is too large. As can be seen from Table 5, compared with APG_CRL, after penalizing the angle, the average time to reach the target point is reduced by 0.19 s. Since APG_CARL avoids large angular changes, it can reduce the oscillation behavior of the robot and improve the efficiency of obstacle avoidance to a certain extent. The action selection of the robot in a certain obstacle avoidance task is shown in Figure 11. Figures 11(a) and 11(b) represent the simulation environment where APG_CRL and APG_CARL are located respectively. The red arrow represents the movement direction of the robot. Figures 11(c) and 11(d) represent the value obtained by the robot after performing the action in the action space. The lighter the color, the higher the value obtained after performing the action. Since APG_RL only considers the comfortable distance of pedestrians, in the situation shown in Figure 11, the robot tends to choose an angular velocity with a greater angular change but a farther distance from pedestrians. While APG_CARL penalizes cases where the angular velocity changes greatly. Under the same circumstances, the optimal selection range of angular velocity is concentrated near the current angular velocity, which will not cause a large jump in the control signal and is more convenient for the control of physical robots in the future.

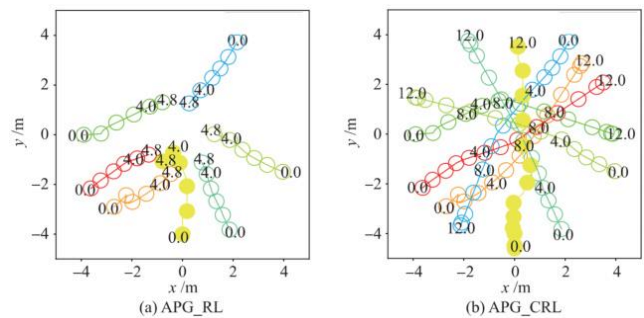

**Fig. 10** Trajectory comparison in a test case among different reward functions

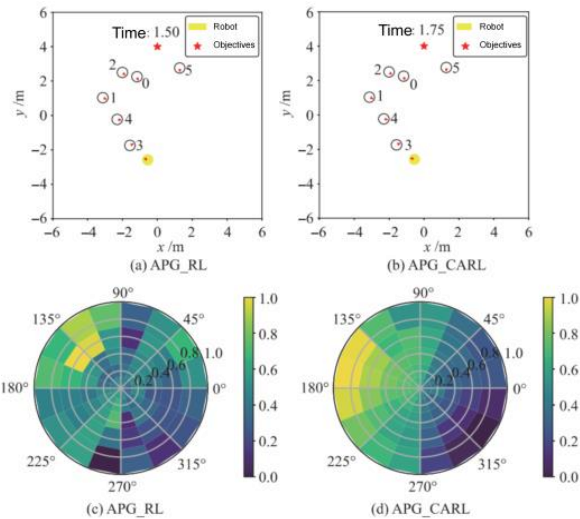

**Fig. 11** The value function result of each feasible action performed by the robot

## 4 Conclusion

The comfortable obstacle avoidance algorithm for mobile robots proposed in this paper first addresses the deficiency that shallow value function networks are difficult to fit complex pedestrian environments. It improves the value function and designs a pedestrian interaction information module. The interaction information between pedestrians is extracted through the angular pedestrian grid, and the temporal features of pedestrian walking trajectories are extracted by using the attention mechanism. Secondly, by modifying the reward function, the pedestrian comfort requirements are introduced into the obstacle avoidance strategy, making the robot's obstacle avoidance strategy more in line with the needs of human-computer interaction. Through comparative analysis of simulation experiments, the algorithm proposed in this paper not only has a good obstacle avoidance success rate and adaptability in complex dynamic environments with dense crowds, but also can meet the requirements of pedestrian comfort.

In the future, this paper will continue to study the influence of different pedestrian comfort zones and dynamic obstacle extraction methods on the performance of the algorithm. In addition, the algorithm proposed in this paper only achieves good results in simulation. Subsequent research will prepare to apply this theoretical method to actual mobile robots, such as NAO and Pepper robots. Through practical applications, the algorithm will be further improved to make it adapt to environmental factors such as inaccurate sensor information in the real environment.